\documentclass[journal]{IEEEtran}
\usepackage{epsfig}
\usepackage[noadjust]{cite}
\usepackage{graphicx}
\graphicspath{{./fig/}}
\usepackage{hyperref}
\usepackage{amsmath, amssymb, amsthm}
\usepackage[caption=false,font=footnotesize]{subfig}

\usepackage{multicol,multirow}
\usepackage{xcolor}
\usepackage{epstopdf}
\usepackage{svg}
\usepackage{svg}

\usepackage{algorithm,algorithmic}

\begin{document}
\title{Bab\_Sak Robotic Intubation System (BRIS): A Learning-Enabled Control Framework for Safe Fiberoptic Endotracheal Intubation}

\author{ Saksham Gupta$^{1}$, Sarthak Mishra$^{1}$,  Arshad Ayub$^{2}$, Kamran Farooque$^{2}$, Spandan Roy$^{1}$, \\ Babita Gupta$^{2}$ 

\thanks{The project was funded by the Indian Council of Medical Research (ICMR) through Extramural Adhoc Grants (Project  I-1248). \href{mailto:drbabitagupta@hotmail.com}{Dr. Babita Gupta} is the principal investigator and corresponding author. Contributions from the first two authors were equal}
\thanks{$^{1}$ The authors are with Robotics Research Center, International Institute of Information Technology Hyderabad (IIIT-H), India. {(email: \{\href{mailto:saksham.g@research.iiit.ac.in}{saksham.g} , sarthak.mishra\}@research.iiit.ac.in,  spandan.roy@iiit.ac.in)}  }
\thanks{$^{2}$ The authors are with J.P.N.Apex Trauma Center, All India Institute of Medical Sciences, New Delhi, India {(email:\{dr.babitagupta, kamran.farooque, dr.arshad\}@aiims.gov.in)} }
\thanks{\textbf{Provisional Patent Number: 202311072640}}}

\maketitle

\begin{abstract}
Endotracheal intubation is a critical yet technically demanding procedure, with failure or improper tube placement leading to severe complications. Existing robotic and teleoperated intubation systems primarily focus on airway navigation and do not provide integrated control of endotracheal tube advancement or objective verification of tube depth relative to the carina. This paper presents the Robotic Intubation System (BRIS), a compact, human-in-the-loop platform designed to assist fiberoptic-guided intubation while enabling real-time, objective depth awareness. BRIS integrates a four-way steerable fiberoptic bronchoscope, an independent endotracheal tube advancement mechanism, and a camera-augmented mouthpiece compatible with standard clinical workflows. A learning-enabled closed-loop control framework leverages real-time shape sensing to map joystick inputs to distal bronchoscope tip motion in Cartesian space, providing stable and intuitive teleoperation under tendon nonlinearities and airway contact. Monocular endoscopic depth estimation is used to classify airway regions and provide interpretable, anatomy-aware guidance for safe tube positioning relative to the carina. The system is validated on high-fidelity airway mannequins under standard and difficult airway configurations, demonstrating reliable navigation and controlled tube placement. These results highlight BRIS as a step toward safer, more consistent, and clinically compatible robotic airway management.
\end{abstract}
{\footnotesize
\noindent Website:  \url{https://sites.google.com/view/robotic-intubation/home}}

\IEEEpeerreviewmaketitle

\section{Introduction}
\begin{figure*}
    \includegraphics[width=\textwidth]{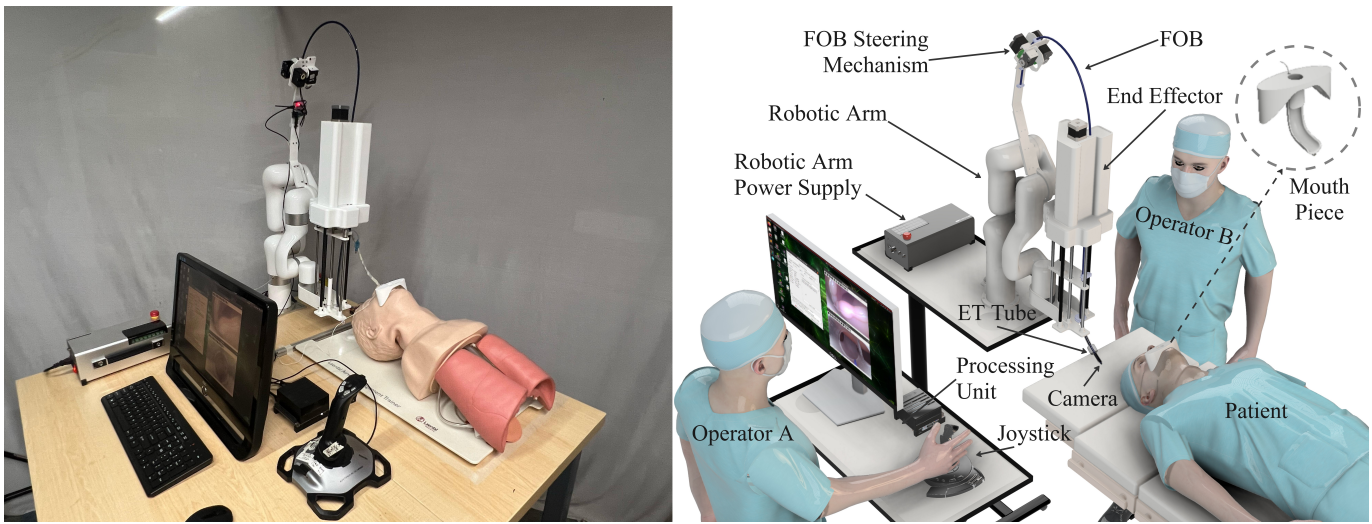}
    \centering
    \caption{ Proposed Robotic Fiberoptic Intubation System featuring the Actual Setup (Left) and its 3D Rendered Model (Right)}
    \label{fig:BRIS}
\end{figure*}

Airway control is imperative in resuscitation, anesthesia, emergency, and critical care, with endotracheal intubation serving as a cornerstone intervention. Despite its routine use, intubation is technically demanding and remains challenging even for experienced clinicians, with failures leading to complications ranging from minor trauma to severe hypoxia, brain damage, and death \cite{cook2011major,cook2011majorpart2,cook2012complications}. These risks were amplified during the COVID-19 pandemic, when intubation was identified as a high-risk aerosol-generating procedure that significantly increased viral exposure to healthcare workers, particularly in emergency settings and while using personal protective equipment \cite{covid}. This underscored the need for safer, more reliable, and remote-capable airway management technologies that reduce direct clinician–patient contact.

In response, robotic and teleoperated approaches have been explored to reduce operator dependence, improve procedural consistency, and enable physical distancing. Joystick-based master–slave systems have demonstrated stable remote navigation of flexible endoscopes with reduced unintentional tissue interactions \cite{deng2024}, while soft and robot-assisted platforms aim to enhance precision and reduce trauma \cite{liu2024soft}. However, existing systems largely focus on navigation assistance and do not provide integrated control of endotracheal tube advancement or objective, real-time verification of tube-to-carina distance.

\subsection{Background and Related works}
Conventional endotracheal intubation is normally accomplished using a direct laryngoscope. The glottic opening is visualized and the endotracheal tube (ET) is inserted through it. However, if visualization of glottic opening is difficult; the chances of failure are high \cite{sun2005glidescope, hansel2022videolaryngoscopy}.

As a result, numerous advancements in airway technology have been made to increase the safety of intubation, even when performed by amateur medical students or by a paramedic in remote locations in inadequate settings. These include video laryngoscopes, intubating supraglottic airway devices, video stylet, flexible bronchoscopes, etc. Flexible fiberoptic bronchoscope (FOB) guided intubation is the gold standard practice for anticipated difficult airways \cite{vasconcelos_pereira_new_2022, grensemann2021tracheal,vlvsdl, boehler2020realiti }. 

FOB-guided intubation can be used for patients with limited mouth opening, cervical instability, oropharyngeal mass, etc. However, it requires training, skill, and expertise. In particular, difficulty is encountered in insertion of fiberscope and rail-roading the ET tube over FOB into the trachea \cite{asai2004fob}. Another major concern while securing an ET tube is its correct and appropriate depth of placement because of the major complications associated with its misplacement. To date, the distance from distal end of the ET tube to the carina is measured by reading the markings on the ET tube \cite{CHONG2006489} or in certain situations by using deep learning models on XRays of the patient \cite{huang2022validation}. However, discrete distance measurement with the help of a combination of FOB-camera and deep learning models has not yet been investigated.

With the advent of deep learning and robotics in every aspect of life, there have also been a few innovations in robotic devices to help aid endotracheal intubation. Although they target different clinical tasks, CathSim’s \cite{cathsim2024} open-source catheter-navigation simulator and a recent robotic-tracheostomy design framework \cite{designReqPDT2024} both show how transparent benchmarking tools and codified safety requirements can shorten the journey from laboratory prototypes to dependable airway robotics. These are supposed to help us decrease the need for expertise, increase the success rate, and overcome the difficulties faced during endotracheal intubation. The Kepler Intubation System (KIS)  \cite{hemmerling2012kepler} is a previously designed robotic intubation system. It has a robotic intubation arm that may be operated with a joystick and a Pentax video laryngoscope. Nevertheless, KIS has limitations such as its large non-portable size and usage of the in-channel video laryngoscope. The In-channel video laryngoscopy necessitates a minimal mouth opening of 1.5- 2 fingers, may incite a laryngoscopy response and cannot be performed in patients with oropharyngeal mass or deformity In addition, KIS necessitates manual insertion of the ET tube into the trachea, which can be challenging because the tube occasionally deviates from the glottic opening and does not always follow the laryngoscope's path \cite{ kriege2017ambu}.

Early robotic platforms such as IntuBot \cite{cheng2019intubot} and the Remote Robot-Assisted Intubation System (RRAIS) \cite{wang2018remote} represent important steps toward automating endotracheal intubation through vision-based navigation. While these systems demonstrate the feasibility of robotic airway access, they primarily focus on guiding the endoscopic view toward the glottis and do not address the broader challenges encountered in clinical environments. In particular, they lack mechanisms for controlled endotracheal tube advancement and objective, real-time verification of the distal tube tip position relative to the carina, a clinically critical parameter for safe tube placement. Similarly, handheld robotic platforms such as REALITI \cite{boehler2020realiti} concentrate on steering assistance to achieve glottic visualization but provide no quantitative feedback on tube depth, limiting their effectiveness in difficult airway scenarios.

More recent work has investigated soft robotic approaches to improve safety and consistency during intubation. Compliant, bendable, or everting structures combined with visual feedback have demonstrated reduced insertion forces and high success rates in simulated and animal airway models \cite{liu2024soft,Haggerty2025SoftIntubation}. However, these systems often depend on external actuation units or custom disposable tubes with embedded actuation, increasing mechanical complexity and limiting portability, reuse, and compatibility with standard clinical workflows. Furthermore, while such approaches facilitate tracheal entry, they do not provide real-time, objective estimation of the tube-to-carina distance required to confirm safe and optimal tube placement.

In parallel, teleoperated robotic platforms employing master–slave architectures and joystick-assisted navigation have been proposed to improve image-guided maneuverability of flexible endoscopes during intubation \cite{deng2024}. These systems enhance operator control and reduce unintentional tissue interactions but typically assume a fixed and stabilized base and focus on local image-space navigation. As a result, they do not address integrated endotracheal tube railroading or quantitative depth verification during placement. Collectively, despite significant progress across autonomous, soft robotic, and teleoperated paradigms, there remains a lack of compact, clinically compatible systems that integrate controlled tube advancement with real-time, objective tube-to-carina depth verification within a human-in-the-loop framework.

This raises a foundational question: can a single, compact, and intelligent robotic system be designed to not only assist with difficult endotracheal intubation, but also objectively measure and verify safe tube depth placement in real time? This paper answers this question by introducing the Robotic Intubation System (BRIS).

\vspace{-10pt}
\subsection{Contributions}
This paper makes the following technical contributions toward safe, clinically compatible robotic airway management within a human-in-the-loop framework. An overview of the integrated system architecture and hardware platform is shown in Figure~\ref{fig:BRIS}.
\begin{itemize}
    \item \textbf{Robotic Intubation Hardware:}  
    Design and fabrication of a compact robotic fiberoptic intubation system comprising a four-way steerable bronchoscope, independent endotracheal tube advancement, and a camera-augmented mouthpiece compatible with standard clinical workflows.

    \item \textbf{Camera-Augmented Mouthpiece Design:}  
    Development of a custom mouthpiece with integrated visual sensing to assist glottic alignment and initial airway access under challenging anatomical conditions.

    \item \textbf{Learning-Enabled Closed-Loop FOB Control:}  
    A learning-based control model that maps joystick inputs to distal FOB tip motion in Cartesian space using real-time shape sensing, enabling closed-loop regulation under tendon nonlinearities and airway contact.

    \item \textbf{Human-in-the-Loop System Integration:}  
    Integration of hardware, perception, and control within a joystick-operated framework that supports intuitive teleoperation with perception-aware guidance.

    \item \textbf{Experimental Validation:}  
    Validation on high-fidelity airway mannequins demonstrating reliable navigation and controlled tube placement across standard and difficult airway scenarios.
\end{itemize}

\section{Hardware Construction}
Our Proposed system's hardware structure, which system consists of 3 components, is explained below:
\subsection{Four-Way Steerable Fiber-Optic Bronchoscope}
\begin{figure}[htbp]
    \includegraphics[width=.4\textwidth]{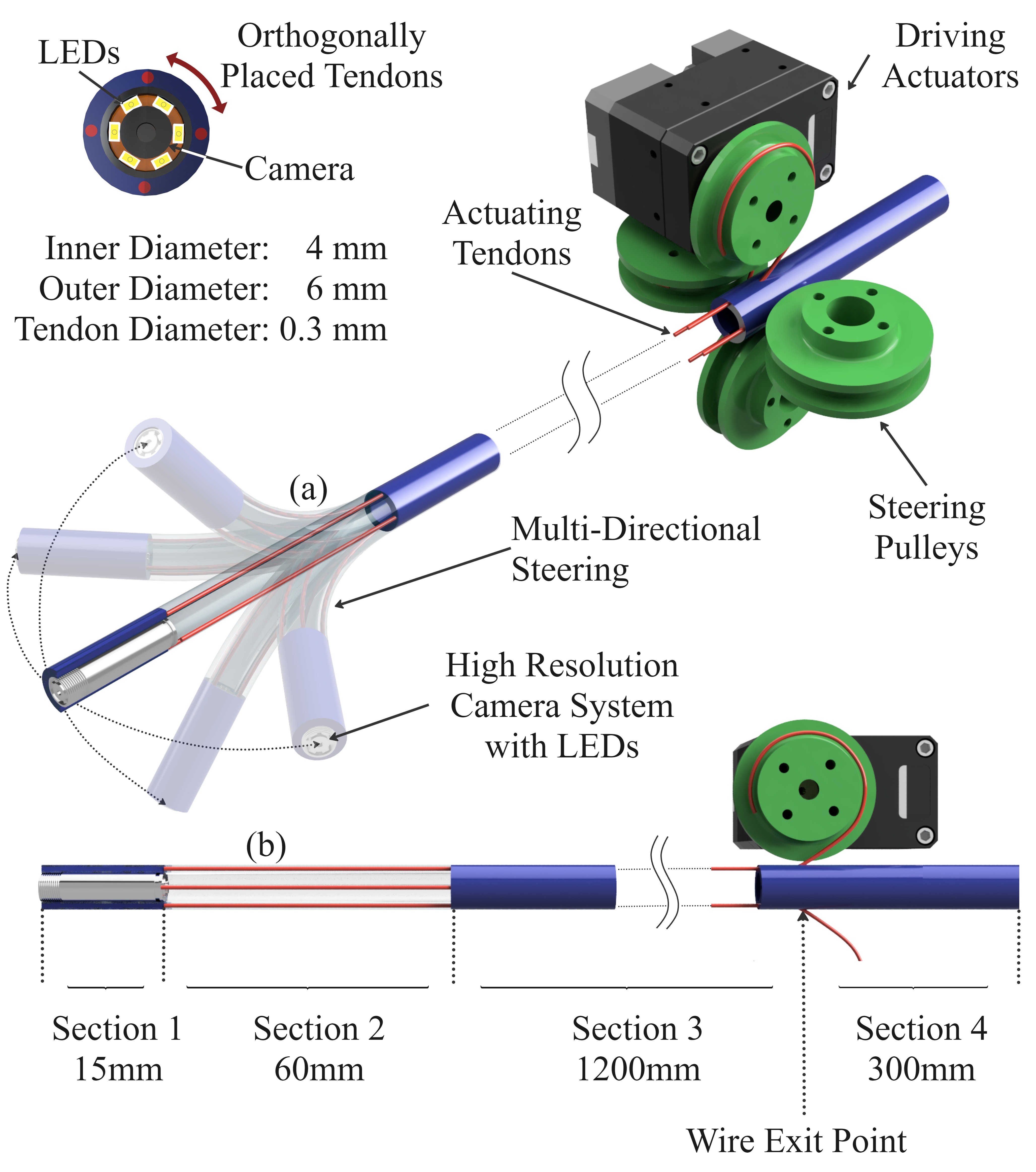}
    \centering
    \caption{Manufactured Four-way Fiberoptic Bronchoscope showcasing Detailed Design and Functional Features.}
    \label{fig:cath}
\end{figure}
The proposed four-way steerable fiber-optic bronchoscope (FOB), shown in Fig.~\ref{fig:cath}, has a total length of $1500$~mm, an outer diameter of $6$~mm, and an inner diameter of $4$~mm, maintained within a $0.1$~mm tolerance. The inner lumen accommodates a distal camera and an embedded fiber-optic shape-sensing (FOSS) sensor, while the outer diameter allows coaxial mounting of standard adult endotracheal tubes ($7.5$--$8.5$~mm). The shaft is fabricated using two Pebax grades: a rigid Pebax~7233 ($M_1$) and a compliant Pebax~3533 ($M_2$), selected to balance axial stiffness and steerability.

The FOB is divided into four functional sections. The distal tip (Section~1) and proximal shaft (Sections~3--4) are composed of the rigid material $M_1$ to protect the camera housing and provide sufficient column strength for ET tube advancement. Section~2 is fabricated from the compliant material $M_2$ and serves as the active articulation zone.

Steering is achieved via four $0.3$~mm stainless steel tendons anchored orthogonally at the junction of Sections~1 and~2 and routed to an external steering module. The tendons are actuated by four Dynamixel XM430-W210~T servo motors arranged orthogonally around the FOB, enabling continuous distal tip orientation control in $\mathbb{S}^2$ through differential tendon actuation.

A fiber-optic shape-sensing (FOSS) sensor (Shape Sensing Company)\footnote{https://shapesensing.com/} is embedded along the FOB lumen, providing real-time reconstruction of the three-dimensional centerline shape. This sensing capability enables direct estimation of shaft configuration and distal tip pose, forming the basis for closed-loop, model-based, and learning-enabled control strategies described in subsequent sections.

\vspace{-10pt}

\subsection{End-Effector and Tube Advancement Mechanism}
\begin{figure*}[htbp]
    \includegraphics[width=0.8\textwidth]{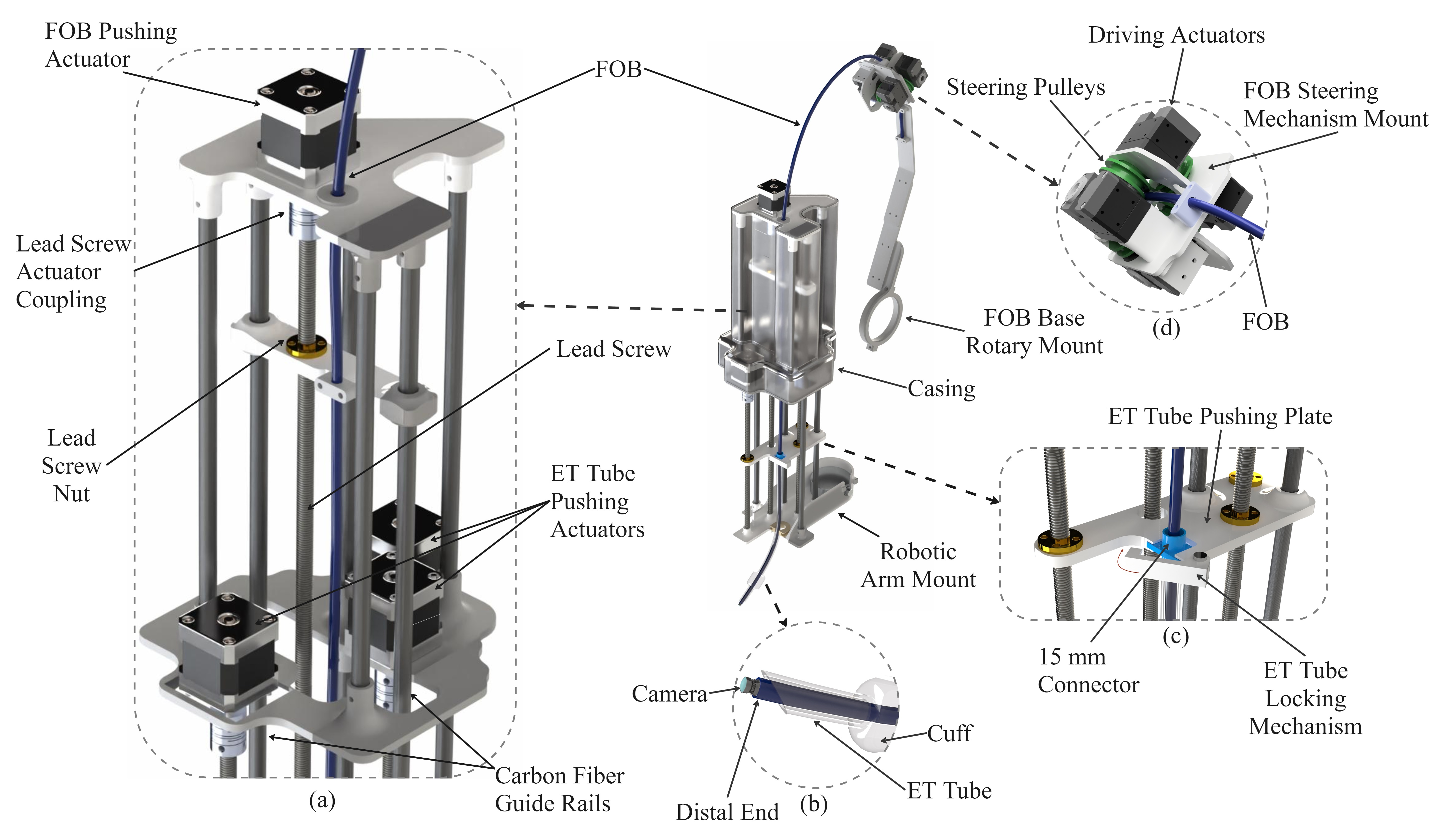}
    \centering
    \caption{Detailed View of the Proposed End Effector for Intubation, Highlighting Its Design and Functional Specifications}
    \label{fig:end effector}
\end{figure*}

The custom end effector assembly, shown in Fig. \ref{fig:end effector}, is designed to interface with the robotic manipulator via a dedicated mounting bracket. To achieve the required independent linear translation of the Endotracheal (ET) tube and the Fiber Optic Bronchoscope (FOB), the system utilizes a lead screw mechanism driven by NEMA 17 stepper motors. The actuation architecture is stratified: the motors driving the ET tube are housed on a central platform, while the FOB actuator is positioned on the superior platform. Structural integrity is maintained by roll-wrapped carbon fiber tubes that connect the mounting plates to the robotic arm interface; these tubes serve a dual function, providing rigid support while simultaneously acting as linear guide rails for the pushing plates that advance the instruments. Regarding modularity, the FOB is strictly secured to ensure intra-operative stability but remains replaceable for sterilization or maintenance between procedures. Conversely, the ET tube interface incorporates a specialized locking mechanism designed to facilitate rapid mounting and dismounting.
\vspace{-10pt}
\subsection{Camera Augmented Mouthpiece}
\begin{figure}
    \includegraphics[width=.5\textwidth]{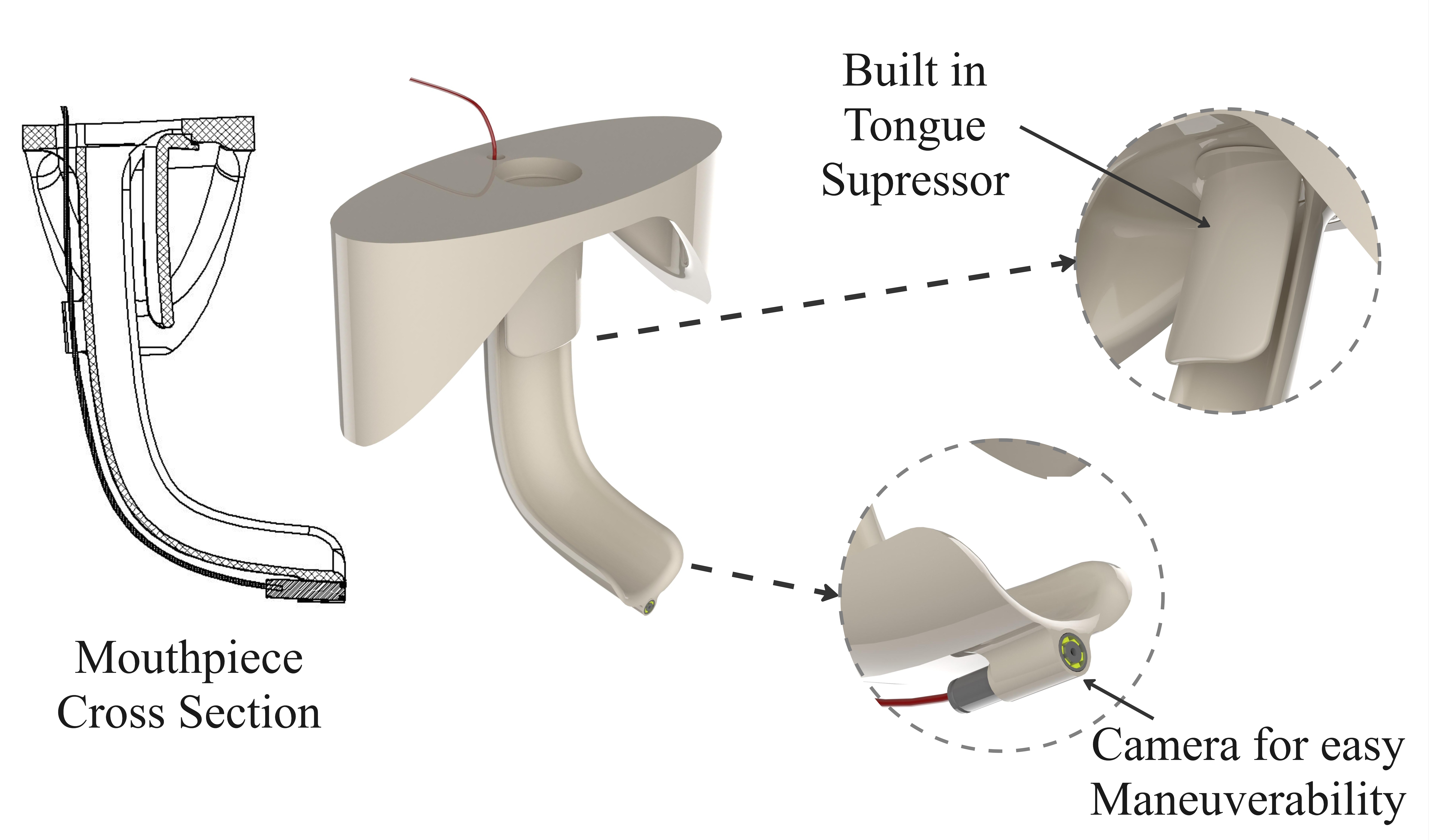}
    \centering
    \caption{Cross section and Design of the manufactured Mouth Piece}
    \label{fig:mouthpiece}
\end{figure}

To facilitate the initial insertion of the ET tube, a custom mouthpiece was developed based on the curvature and geometry of the standard Macintosh laryngoscope blade. This component incorporates an integrated camera module that provides a continuous visual field of the glottis. This visual feedback loop is critical for the robotic intubation workflow, as it offers the operator real-time situational awareness to adjust the FOB trajectory and ensure precise instrument placement within the airway. Fig.~\ref{fig:mouthpiece} shows the cross-sectional design of the camera-augmented mouthpiece, highlighting its geometry and integrated visual sensing channel.

\section{Software Design}
\subsection{Intelligent Depth Estimation and Anatomy-Aware Guidance}

Safe navigation and placement of the endotracheal (ET) tube require accurate depth perception and continuous alignment with the tracheal lumen. We address these requirements through an integrated perception framework that combines airway zone classification with passive, anatomy-aware visual guidance using monocular endoscopic imaging.

The airway is partitioned into three clinically relevant zones along the tracheobronchial axis:
\begin{itemize}
    \item \textbf{Zone I (Supraglottic and Proximal Tracheal Region):} Includes the laryngeal inlet, vocal cords, and proximal trachea distal to the glottis.
    \item \textbf{Zone II (Optimal Tracheal Placement Region):} Corresponds to the mid-tracheal segment and represents the recommended safety window for ET tube placement.
    \item \textbf{Zone III (Distal Tracheal and Carinal Region):} Defined by proximity to the carina ($<1$~cm) or entry into a mainstem bronchus, indicating elevated risk of endobronchial intubation.
\end{itemize}

\begin{figure}[b]
    \includegraphics[width=.5\textwidth]{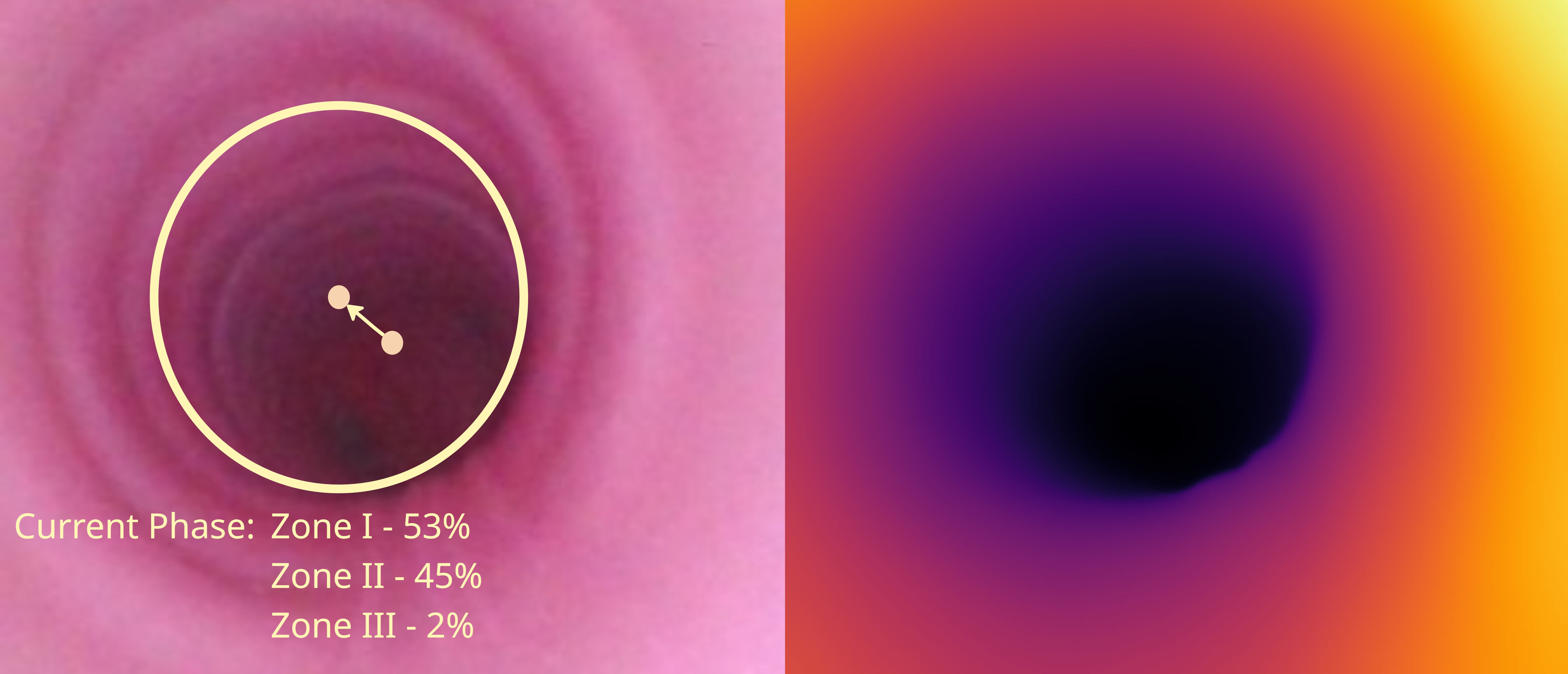}
    \centering
    \vspace{-7mm}
    \caption{Anatomy-aware visual guidance using monocular depth estimation for airway zone 
classification and passive lumen-alignment during fiberoptic intubation.}
    \label{fig:vis_guidance}
\end{figure}
Fig.~\ref{fig:vis_guidance} illustrates the anatomy-aware visual guidance pipeline, including airway zone classification from monocular depth estimation and the passive lumen-alignment overlay used during navigation.
Dense depth perception is obtained by directly deploying the pre-trained \textit{Depth Anything} foundation model~\cite{yang2024depthanything}, exploiting its zero-shot generalization without medical-domain fine-tuning. The model processes the live endoscopic video stream to generate real-time depth maps, which are analyzed to estimate the relative distance between the FOB tip and the carina. This estimate is used to assign the current state to one of the three airway zones and to provide a simplified guidance cue to the operator: \textit{Advance}, \textit{Maintain Position}, or \textit{Withdraw}.

In parallel, the same depth maps are used to provide continuous alignment feedback during navigation. Rather than employing fully autonomous visual servoing, we adopt a \emph{passive visual servoing} approach that augments operator perception while preserving full manual control. The system identifies the image-space location of maximum estimated depth, corresponding to the tracheal lumen center $C_{lumen}$. Let $C_{cam}$ denote the optical center of the endoscopic camera; the misalignment vector is computed as
\begin{equation}
    \vec{v} = C_{lumen} - C_{cam}.
\end{equation}

This vector is rendered as a dynamic overlay on the endoscopic display, indicating the direction and magnitude of off-axis deviation. By minimizing $|\vec{v}|$ through joystick inputs, the operator maintains coaxial alignment of the FOB with the tracheal lumen, reducing wall contact and mitigating mucosal trauma during advancement.

Overall, the proposed framework unifies depth-based airway state recognition and interpretable visual guidance within a human-in-the-loop control paradigm suitable for safety-critical airway management.

\subsection{Human--Robot Interface and Software--Hardware Integration}
\begin{figure}[htbp]
    \includegraphics[width =.5 \textwidth]{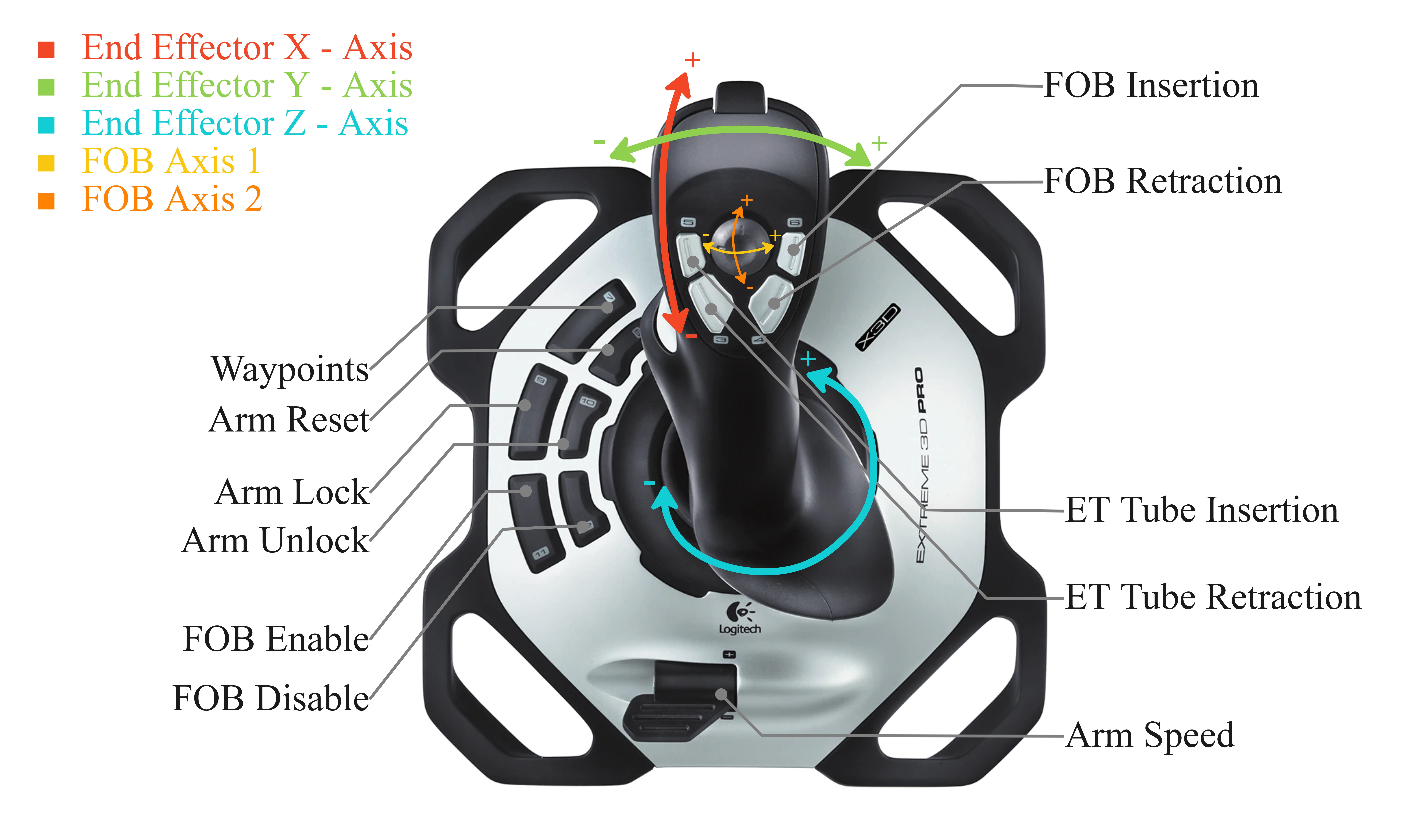}
    \centering
    \caption{Joystick Controller Mapping - An Overview of Control Assignments and Interaction Mechanisms}
    \label{fig:joystick}
\end{figure}
Fig.~\ref{fig:joystick} provides an overview of the joystick-based human--robot interface used to teleoperate the Robotic Intubation System (BRIS), including the mapping between controller inputs and instrument-level control actions. A joystick-based interface enables intuitive teleoperation of the Robotic Intubation System (BRIS). As the robotic arm SDK does not natively support joystick inputs, a custom mapping layer was implemented using the Cartesian velocity control mode. Joystick axis deflections are normalized to $[-1,1]$ and mapped to the end-effector velocity vector $\left[v_x, v_y, v_z\right]$, while a sliding throttle acts as a scalar gain to dynamically limit execution speed for safe and precise manipulation.

Fine instrument control is provided via the controller’s \textit{hat switch}, which independently actuates the four steering directions of the fiber optic bronchoscope (FOB). Dedicated buttons trigger auxiliary actions, including endotracheal tube retraction and automated motion to pre-defined robotic arm waypoints.

High-level control and hardware abstraction are centralized on an NVIDIA Jetson AGX Xavier running Linux and ROS. Custom ROS nodes translate user inputs into hardware commands. The robotic manipulator communicates over Ethernet, while Dynamixel servos and camera modules use serial links. Linear insertion and retraction are driven by stepper motors controlled through a microcontroller-based motor driver interfaced via USB, resulting in a modular and extensible control architecture.

\section{Learning-Based Control Framework for the Steerable Fiberoptic Intubation Robot}
Robotic control of tendon-driven flexible instruments is challenging due to frictional hysteresis, nonlinear tendon coupling, and the lack of an analytical model relating actuator torques to continuum deformation. In BRIS, these effects are intensified by airway contact, anatomical variability, and the requirement for precise 3-D tip control. To achieve reliable closed-loop navigation, we adopt a data-driven dynamics \cite{saviolo2023learning} model that learns the mapping from actuator torques and backbone shape to future tip state.
\subsection{Learning-Based Dynamics Model}

Accurate prediction of the bronchoscope tip motion is essential for safe navigation in the airway, yet tendon-driven continuum robots exhibit nonlinear coupling, frictional hysteresis, and deformation patterns that cannot be captured by tip position alone. The ShapeSensing\texttrademark{} backbone reconstruction provides rich information about curvature, torsion, and internal loading, but this high-dimensional signal must be compressed into a tractable state representation for control. To address this, we formulate a data-driven forward dynamics model that integrates tip state, actuator torques, and a latent encoding of the backbone shape.

Let $p_t \in \mathbb{R}^3$ denote the tip position and compute the velocity by finite differences, ${v}_t = \frac{{p}_t - {p}_{t-1}}{\Delta t},$
yielding the system state
\begin{equation}
{x}_t =
\begin{bmatrix}
{p}_t &
{v}_t
\end{bmatrix} ^{\top}
\in \mathbb{R}^6.
\end{equation}
The ShapeSensing\texttrademark{} module returns the ordered backbone point cloud
$
\mathbf{S}_t = \{p^{(1)}_t,\dots,p^{(N)}_t\},
$
capturing the distributed deformation along the flexible segment.  
The control input is the vector of tendon and insertion torques,
\begin{equation}
{u}_t =
[\tau^{(1)}_t,\tau^{(2)}_t,\tau^{(3)}_t,\tau^{(4)}_t,\tau^{(5)}_t]^\top .
\end{equation}

A Temporal Convolutional Network (TCN) encoder $E_\phi$ processes a short history of length $L$ of backbone measurements to extract a compact latent representation,
\begin{equation}
{z}_t = E_\phi(\mathbf{S}_{t-L:t}).
\end{equation}
Encoding the deformation sequence, rather than a single frame, provides three advantages: (i) it captures curvature and torsional modes governing the robot's instantaneous response; (ii) it embeds short-term deformation history that reflects frictional loading and hysteresis; and (iii) it suppresses measurement noise and yields a temporally consistent internal state.

\begin{figure*}[h]
    \includegraphics[width = \textwidth]{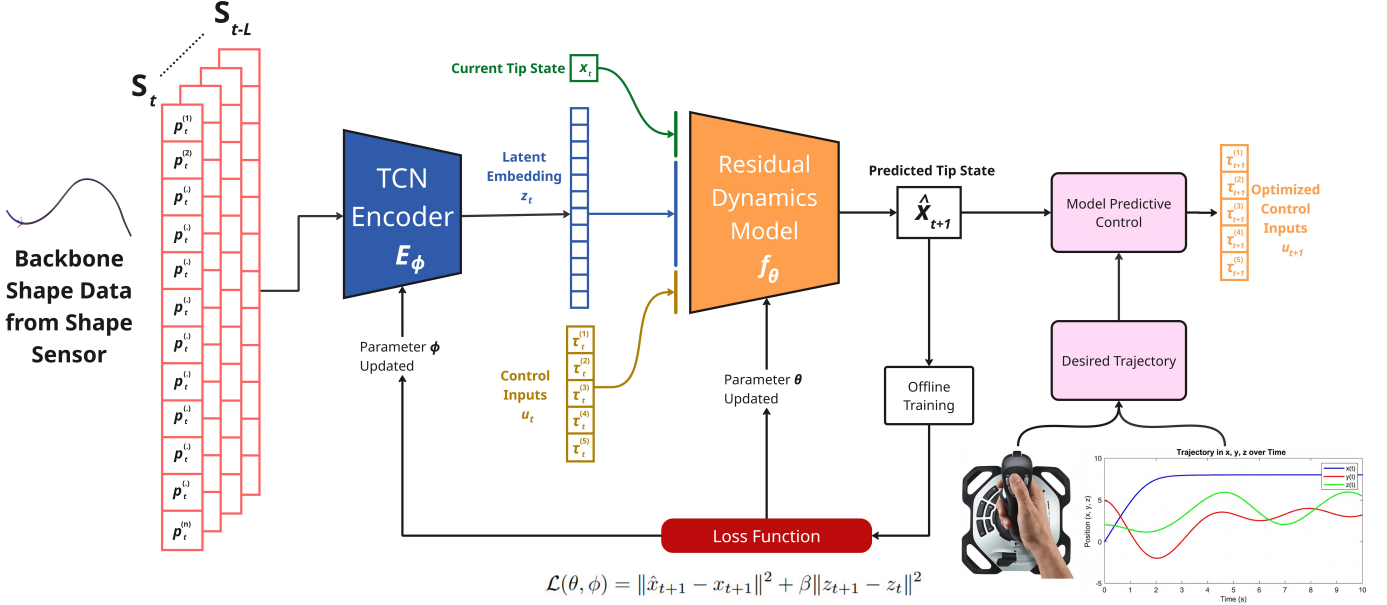}
    \centering
    \caption{Learning-based dynamics and control architecture for the steerable fiberoptic intubation robot. 
    A short temporal window of backbone shape reconstructions from the ShapeSensing\texttrademark{} module is encoded by a TCN-based latent encoder $E_\phi$ into a compact deformation state $z_t$. 
    This latent representation, together with the current tip state $x_t$ and actuator torque inputs $u_t$, is provided to a residual neural dynamics model $f_\theta$ to predict the next tip state $\hat{x}_{t+1}$. 
    The model is trained end-to-end using tip-state prediction and latent temporal smoothness losses. 
    During execution, the learned dynamics are embedded within a nonlinear model predictive control (MPC) framework to compute optimal control inputs for closed-loop trajectory tracking.}
    \label{fig:learning_model}
\end{figure*}

Fig.~\ref{fig:learning_model} provides an overview of the proposed learning-based dynamics and control architecture, showing how backbone shape information, actuator inputs, and tip-state feedback are combined for closed-loop prediction and control.

The forward dynamics are modeled as:
\begin{equation}
\hat{{x}}_{t+1} = f_\theta({x}_t, {z}_t, {u}_t),
\end{equation}
where $f_\theta$ is a lightweight residual neural network.  
By conditioning on both the tip state and the latent backbone representation, the model learns the implicit relationships between actuator inputs, stored tendon tension, contact-induced deformation, and resulting tip motion—effects that are analytically intractable in continuum mechanics.

Since ${z}_t$ is a learned latent variable, training relies on tip-state prediction and temporal regularization:
\begin{equation}
\mathcal{L}(\theta,\phi)
=
\|\hat{{x}}_{t+1} - {x}_{t+1}\|^2
+ \beta \|{z}_{t+1} - {z}_{t}\|^2
\end{equation}
This objective encourages accurate prediction of the next tip position and velocity while promoting smooth latent transitions that reflect physically plausible backbone evolution.  
Gradients propagate through both the dynamics model parameters $\theta$ and the encoder parameters $\phi$, enabling fully end-to-end optimization without auxiliary losses or staged training.  
The resulting dynamics model leverages high-dimensional shape information to provide reliable, temporally coherent predictions under anatomical variability, compliant wall interactions, and the nonlinear behavior inherent to tendon-driven continuum instruments.

\subsection{Model Predictive Control}
Finally, to achieve robust closed-loop trajectory tracking, we integrate our adaptive dynamics model into a model predictive control (MPC) framework that explicitly accounts for predictive uncertainty. During execution, control inputs are computed in real time using a nonlinear MPC controller. Specifically, at each time step $k$, the controller solves a finite-horizon optimal control problem (OCP) of the form:
\begin{align}
\min_{\{x_{k+i}, u_{k+i}\}} \quad  J(x_{k:k+M}, u_{k:k+M-1}) &:= \sum_{i=0}^{M-1} J_s(x_{k+i}, u_{k+i}) \nonumber \\
&+ J_f(x_{k+M})  \\
\text{s.t.} \quad  x_{k+i+1} = f(x_{k+i}, z_{k+i}, u_{k+i}), &\quad x_{k+i} \in \mathcal{X},\ u_{k+i} \in \mathcal{U}, \nonumber \\
 x_0 = x_k, \quad x_M \in \mathcal{X}_f&, \quad \forall i = 0,\dots,N-1 \nonumber
\end{align}
where \( J_s(x_{k+i}, u_{k+i}) = \|x_{k+i} - x_{k+i}^r\|_Q^2 + \|u_{k+i} - u_{k+i}^r\|_R^2 \) is the stage cost, and \( J_f(x_{k+M}) = \|x_{k+M} - x_{k+M}^r\|_P^2 \) is the terminal cost. The reference trajectories \( x_{k+i}^r \), \( u_{k+i}^r \) are given, and \( Q \), \( R \), and \( P \) are positive semi-definite weighting matrices. At each control step $k$, the first control input \( u_{k}^\star \) is executed, and the OCP is re-solved in a receding horizon manner using the updated state estimate.

\section{Intubation Procedure}

The robotic intubation workflow is designed as a supervised, human-in-the-loop procedure involving two personnel: a primary operator (Operator~A) and a clinical assistant (Operator~B). The patient is assumed to be positioned anterior to the system with an unobstructed access path, while the precise pose of the oral cavity and airway anatomy is initially unknown. Operator~B prepares the instrumentation by mounting the endotracheal (ET) tube coaxially over the fiber-optic bronchoscope (FOB), securing the locking adapter, and lubricating the assembly. The custom mouthpiece is then inserted into the oral cavity, and Operator~B maintains continuous access to the hardware emergency stop throughout the procedure.

The approach phase combines autonomous positioning with operator supervision. Operator~A initiates motion of the robotic arm through a sequence of pre-defined waypoints that safely guide the system to a region proximal to the patient’s face. This semi-autonomous positioning reduces manual workload while accommodating inter-patient anatomical variability. Upon reaching the final waypoint above the oral aperture, Operator~A transitions to direct joystick control to perform fine alignment and insert the distal tip of the ET tube--FOB assembly into the mouthpiece channel.

Airway navigation is performed under continuous endoscopic visualization from the FOB-mounted camera. The operator steers the FOB using joystick inputs while receiving real-time, anatomy-aware guidance from the perception and control framework. Dense depth estimation provides a coarse localization of the FOB relative to the airway anatomy and assigns the current state to clinically relevant airway zones. In parallel, a passive visual servoing overlay indicates off-axis deviation from the tracheal lumen center, enabling the operator to maintain coaxial alignment during advancement.

As the FOB progresses beyond the vocal cords, shape-sensing feedback continuously reconstructs the three-dimensional configuration of the flexible shaft. This information is internally leveraged by the learning-based dynamics model and model predictive controller to generate smooth, constraint-aware motion responses, while preserving full operator authority. The FOB is advanced until the carina is centered in the visual field and the depth estimation module classifies the tip location within the optimal placement zone (approximately $2$~cm proximal to the carina).

Once the target depth is reached, the FOB position is held to serve as a stable guide. The ET tube is then mechanically advanced (railroaded) over the FOB until its distal tip becomes visible in the endoscopic view, confirming correct placement. Finally, the FOB is withdrawn, the robotic system is retracted from the patient, and the airway is secured.

\section{Experimentation \& Results}

\subsection{Experimental Protocol}
Due to the invasive nature of endotracheal intubation, validation was conducted on high-fidelity airway training mannequins (Laerdal Medical and Ambu), which replicate realistic airway geometry, tissue compliance, and insertion resistance. This approach enables controlled, repeatable evaluation while adhering to ethical and safety constraints.

Experiments were performed under two clinically representative configurations:
\begin{itemize}
    \item \textbf{Standard airway:} Unobstructed visualization of the glottis (Cormack--Lehane Grade~I).
    \item \textbf{Constrained airway:} Reduced mouth opening, limited cervical mobility, and tongue displacement, yielding a narrower workspace and reduced visibility.
\end{itemize}

Across all trials, performance was assessed in terms of successful tracheal placement, final tube position relative to the carina, and operator interaction characteristics.
\vspace{-10pt}
\subsection{Implementation Details}

The ShapeSensing\texttrademark{} module reconstructs the FOB backbone as an ordered point cloud of $N=48$ points along the flexible segment. A temporal window of length $L=6$ (approximately $120$~ms) is processed by a TCN-based encoder to extract a 16-dimensional latent deformation state. The system state comprises tip position and velocity in $\mathbb{R}^6$, while the control input consists of four tendon torques and one insertion torque. Training data are collected via joystick teleoperation and include straight insertions, sharp bends, S-curves, and contact-rich airway interactions, yielding approximately $6.2\times10^4$ state–transition samples. The latent dynamics model is trained end-to-end using the Adam optimizer with a learning rate of $1\times10^{-3}$, batch size $256$, and latent smoothness weight $\beta=0.1$, converging within 120 epochs. During execution, the learned dynamics are embedded in a nonlinear MPC operating at 50~Hz with a prediction horizon of $M=10$, quadratic state and input costs, and anatomically motivated safety constraints, enabling real-time control with an average computation time below 12~ms per step.

\subsection{Task Success and Placement Reliability}
BRIS was evaluated over 48 complete intubation trials (24 standard and 24 constrained airway configurations). A trial was considered successful if the endotracheal (ET) tube was advanced into the trachea and stabilized within the recommended mid-tracheal placement region (Zone~II).

The system achieved a 100\% success rate across both standard and constrained airway trials. In constrained configurations, transient proximity to the carina (Zone~III) was detected in a small subset of trials; in all such cases, the depth estimation module triggered controlled withdrawal prior to tube advancement, resulting in safe final placement. No endobronchial insertions, esophageal misplacements, or unrecognized failures were observed in any trial.

\subsection{Tube-to-Carina Depth Estimation Accuracy}
Monocular depth-based placement accuracy was quantified by comparing the estimated ET tube tip--to--carina distance with ground-truth measurements derived from known mannequin geometry. Across all trials, BRIS achieved a mean absolute error of $2.4 \pm 1.1$~mm. Standard airway configurations yielded lower error ($2.0 \pm 0.9$~mm), while constrained airways exhibited slightly higher error ($2.8 \pm 1.2$~mm), primarily due to partial occlusions and off-axis viewing.

It is important to note that the reported error reflects relative depth estimation accuracy sufficient for reliable airway zone classification, rather than absolute metric reconstruction at sub-millimeter precision.

Overall, 98\% of trials resulted in final tube placement within $\pm 20$~mm of the target carinal offset, a range widely accepted as clinically safe. These results indicate that the proposed monocular framework provides precise and reliable depth feedback without requiring dedicated depth sensors, structured illumination, or patient-specific calibration.
\vspace{-10pt}

\subsection{Control Stability and Visual Guidance Ablation}
Operator joystick inputs were analyzed to assess the effect of learning-based control on distal navigation. With the full BRIS control stack enabled, joystick input variance during fine positioning distal to the vocal cords was reduced by 46\% relative to baseline teleoperation without shape-aware dynamics, reflecting improved predictability and reduced corrective effort during contact-rich navigation.

A focused ablation study (16 trials) examined the contribution of anatomy-aware visual guidance. Disabling the passive lumen-alignment vector and zone overlays resulted in a 52\% increase in distal wall contacts and a 35\% increase in corrective withdrawals. These findings confirm that the visual guidance cues substantially improve navigation efficiency and placement consistency while maintaining full operator authority and interpretability.

\vspace{-10pt}

\section{Conclusion}

This work introduced the Robotic Intubation System (BRIS), a compact robotic platform that addresses two persistent limitations in robotic airway management: the lack of integrated endotracheal tube control and the absence of objective, real-time tube depth verification. By combining a four-way steerable fiberoptic bronchoscope, independent tube advancement, and a camera-augmented mouthpiece, BRIS enables controlled, workflow-compatible fiberoptic intubation within a human-in-the-loop framework. A learning-enabled closed-loop control strategy leverages real-time shape sensing to map joystick inputs to Cartesian distal tip motion, providing stable and intuitive operation under tendon nonlinearities and airway contact. In parallel, monocular endoscopic depth estimation supplies interpretable, anatomy-aware guidance for positioning the tube relative to the carina. Experimental results on high-fidelity airway mannequins demonstrate that BRIS can reliably support navigation and controlled tube placement across standard and difficult airway scenarios. These findings position BRIS as a step toward clinically practical robotic intubation systems that augment operator capability while improving procedural consistency and safety.

\bibliographystyle{IEEEtran}
\bibliography{root}

\end{document}